\documentclass[sigconf]{acmart}

\copyrightyear{2026}
\acmYear{2026}
\setcopyright{cc}
\setcctype{by}
\acmConference[MM '26]{Proceedings of the 34th ACM International Conference on Multimedia}{November 10--14, 2026}{Rio de Janeiro, Brazil}
\acmBooktitle{Proceedings of the 34th ACM International Conference on Multimedia (MM '26), November 10--14, 2026, Rio de Janeiro, Brazil}
\acmDOI{10.1145/3767308.3836441}
\acmISBN{979-8-4007-2213-4/2026/11}

\AtBeginDocument{%
  }

\newcommand{\cmark}{\checkmark}
\begin{document}

\title{Talking to Me or Someone Else? Rethinking Talk-to-Me Detection in Egocentric Videos}

\author{Feiyu Du}
\affiliation{%
  \institution{The University of Texas at Dallas}
  \city{Richardson}
  \country{United States}
}
\email{feiyu.du@utdallas.edu}

\author{Xi He}
\affiliation{%
  \institution{The University of Texas at Dallas}
  \city{Richardson}
  \country{United States}
}
\email{xi.he@utdallas.edu}

\author{Jia Li}
\affiliation{%
  \institution{The University of Texas at Dallas}
  \city{Richardson}
  \country{United States}
}
\email{jia.li@utdallas.edu}

\author{Yapeng Tian}
\authornote{Corresponding author.}
\affiliation{%
  \institution{The University of Texas at Dallas}
  \city{Richardson}
  \country{United States}
}
\email{yapeng.tian@utdallas.edu}

\author{Weili Wu}
\affiliation{%
  \institution{The University of Texas at Dallas}
  \city{Richardson}
  \country{United States}
}
\email{weiliwu@utdallas.edu}

\renewcommand{\shortauthors}{Feiyu Du, Xi He, Jia Li, Yapeng Tian, and Weili Wu}


\begin{abstract}
Online understanding of who is talking to the camera wearer is a key capability for egocentric social interaction. However, existing talk-to-me (TTM) studies are commonly formulated as offline clip-level recognition, which is poorly aligned with online interaction and overlooks the diverse non-TTM speaking states that naturally arise in egocentric videos. In this paper, we revisit this problem by reformulating it as an online, frame-level prediction task. Instead of treating TTM as a binary problem against a single negative class, we model it in the presence of diverse and previously underexplored non-TTM states, such as talking-to-others, self-talking, and background conditions. To support this new formulation, we construct an Online TTM Dataset consisting of 406 egocentric video clips with approximately 900K annotated frames, each labeled with frame-level social interaction categories (e.g., background, TTM, talking-to-others, self-talking), by extending the Ego4D social interaction benchmark. In this benchmark, we evaluate five adapted baselines and develop a new model that integrates social cues across modalities. Experimental results show that our multimodal model, which jointly leverages audio, visual, and speech-semantic cues, achieves 75.5\% frame-level F1 on TTM, outperforming strong baselines and enabling a systematic analysis of how different speaking states affect TTM recognition.

\end{abstract}

\ccsdesc[500]{Information systems~Multimedia information systems}

\keywords{Egocentric Video Understanding, Talk-to-me Recognition, Multimodal Learning}

\maketitle

\begin{figure*}[t]
  \centering
  \includegraphics[width=0.95\linewidth]{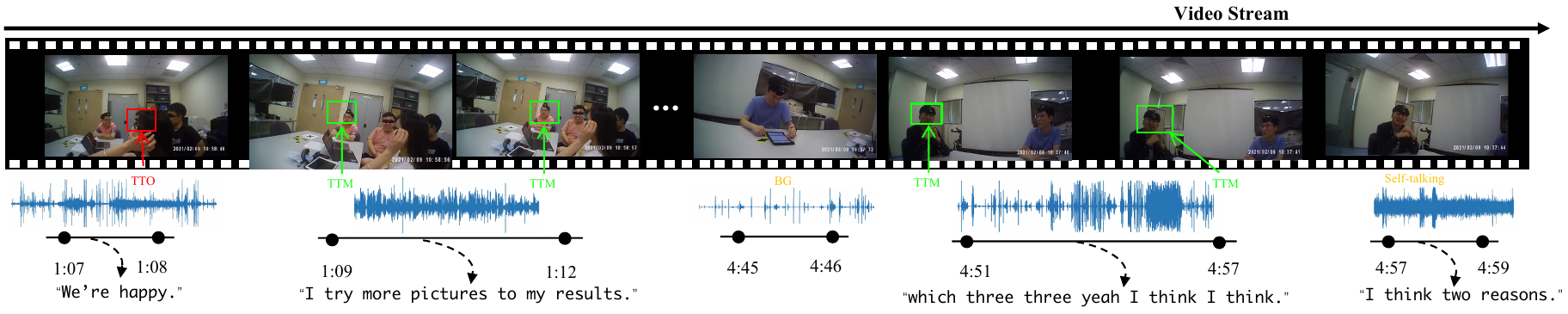}
  \caption{We introduce the Online TTM Dataset, a multimodal benchmark for online who-is-talking-to-me perception in realistic first-person scenarios. It temporally grounds social speaking states in continuous egocentric video streams while integrating visual, audio, and speech information. Within such a continuous first-person stream, the current social speaking state may shift over short temporal windows from background to talking-to-others, talking-to-me, or self-talking.}
  \Description{An overview figure showing a continuous egocentric video stream annotated with four social speaking states over time. The visual layout highlights how the state of interaction changes within a first-person scene, including background, talking-to-others, talking-to-me, and self-talking. The figure emphasizes temporal transitions between these states and illustrates that wearer-directed speech appears as only one part of a richer stream of multimodal social activity.}
  \label{fig:intro_ttm}
\end{figure*}
\section{Introduction}
Understanding who is talking to the camera wearer is a fundamental capability for egocentric social perception systems~\cite{jiang2022egodmcavasl, ryan2023egocentric, clarke2023egodeit, clarke2025egoscan, plizzari2024egovision}. A wearable assistant, an augmented reality interface, or a human-centered robot should not only detect the presence of speech, but also determine whether a visible person is actually addressing the wearer at the current moment. This capability is central to a wide range of interaction-driven applications, such as deciding when a system should respond, identifying the relevant speaker in a multi-person scene, and distinguishing direct interaction from background social activity~\cite{dhaussy2023avsdhri, xu2022avaavd}.

Despite its importance, the talk-to-me (TTM) problem in egocentric video~\cite{kong2024long} has so far been studied primarily under an offline clip-level formulation. In this setting, each video segment is assigned a single label representing its speaking state. While such a design is reasonable for offline recognition, it is not well aligned with online egocentric interaction. In natural social scenes, speaking states often evolve within short temporal windows: a person may first talk to someone else and then turn to the wearer; the wearer may begin speaking; and background music or environmental noise may coexist with socially relevant cues~\cite{tao2021talknet, wang2024loconet}. Consequently, the prevailing clip-level formulation makes it difficult to capture these state transitions and to study how they influence TTM recognition at frame-level granularity~\cite{alcazar2020asc, alcazar2021maas, min2022spell}.

A further limitation lies in how the problem is commonly formulated. Existing TTM settings are often reduced to a binary distinction between \emph{talking-to-me} and \emph{not talking-to-me}~\cite{grauman2022ego4d}. However, the negative side of this formulation is far from uniform. In egocentric social scenes, non-TTM may include background or non-interactive conditions, self-talking by the wearer, and talking-to-others. These cases differ substantially in both social meaning and recognition difficulty. Background is often relatively easy to reject, self-talking places the wearer in the role of speaker, while talking-to-others contains genuine speech and visible facial motion but with a different addressee. Merging all of them into a single negative class hides these differences and makes it difficult to understand what actually causes TTM errors~\cite{zhang2021unicon}. Fig.~\ref{fig:intro_ttm} illustrates how the social speaking state changes over time in a continuous first-person video stream.

Motivated by these issues, we revisit TTM from two aspects. First, rather than treating it as an offline clip-level recognition problem, we reformulate it as an \emph{online frame-level prediction} task, where the model must decide at each moment whether a visible person is speaking to the wearer using only the current and past observations. Second, rather than grouping all non-TTM cases into a single class, we explicitly preserve their differences and study TTM in the presence of multiple socially meaningful speaking states. To support this setting, we construct the Online TTM Dataset, a curated benchmark with approximately 900K frames derived from the Ego4D Social Dataset. Specifically, we reconstruct frame-level supervision for online prediction and re-sample the data to mitigate the substantial class imbalance in the original distribution. During this process, we preserve temporal context around TTM events, continuity within segments, and informative state boundaries, so that the benchmark contains sufficient cross-state variation to support learning transitions between speaking states~\cite{guermal2024joadaa}. We further clean the data by removing missing-frame cases, annotation coverage issues, and samples with prolonged absence of valid face boxes, including consecutive black frames.

On top of this benchmark, we adapt five representative baselines and develop an online multimodal temporal model for TTM recognition. Unlike offline formulations, the model operates in a streaming manner and predicts the current speaking state using only the current frame and recent history, without access to future frames~\cite{kopuklu2021threestage, liao2023lightasd}. To resolve socially confusable cases, the model explicitly integrates fine-grained facial cues, including head-pose-aware visual representations and mouth-motion dynamics~\cite{braga2024sslasd, jung2024talknce}. We further introduce wearer-centered geometric cues to explicitly encode the spatial importance of each face in multi-face scenes, including normalized position, scale, and aspect ratio. These cues provide a strong prior for identifying which face is more likely to be close to the camera, occupy the wearer’s visual focus, and therefore serve as the current interaction target. In addition, the model combines conventional acoustic features with Whisper-derived speech semantics, enabling it to jointly reason about speaking activity, interaction target, and semantic context. By augmenting audio with text-semantic information extracted from speech, we further examine whether richer semantic cues are more effective than acoustic features alone for distinguishing TTM from socially similar speaking states.

Experiments on the Online TTM Dataset show that our proposed online multimodal model achieves 75.5\% frame-level F1 on TTM and consistently outperforms strong baselines across frame-level F1 and segment-level mAP. Beyond overall performance, the proposed benchmark also enables a more comprehensive analysis of the problem. We evaluate the effect of enhancing acoustic features with Whisper-derived speech semantics, and conduct attribution analysis to examine how different speaking states influence TTM recognition. These results show that semantically enhanced audio representations are substantially more effective than conventional acoustic features alone, while four-state modeling provides a more informative formulation for understanding TTM errors and socially confusable cases. 
Our contributions are summarized as follows: 1) We revisit egocentric TTM from two perspectives: we reformulate it from offline clip-level recognition to online frame-level prediction, and replace the simplified TTM and non-TTM setting with a socially structured formulation that better reflects realistic egocentric interaction.
2) We build the Online TTM Dataset, a curated benchmark with approximately 900K frames derived from the Ego4D Social Dataset by reconstructing frame-level supervision, mitigating severe class imbalance through temporally coherent re-sampling, and cleaning noisy cases such as missing frames, annotation coverage issues, and prolonged absence of valid face boxes.
3) We adapt five representative baselines and develop an online multimodal temporal model that integrates fine-grained facial cues, wearer-centered geometry, acoustic features, and Whisper-derived speech semantics, achieving 75.5\% frame-level F1 on TTM while enabling systematic analysis of the effect of different speaking states on recognition.

\begin{figure*}[t]
    \centering

    \begin{minipage}{0.8\textwidth}
        \centering

        \includegraphics[width=\linewidth]{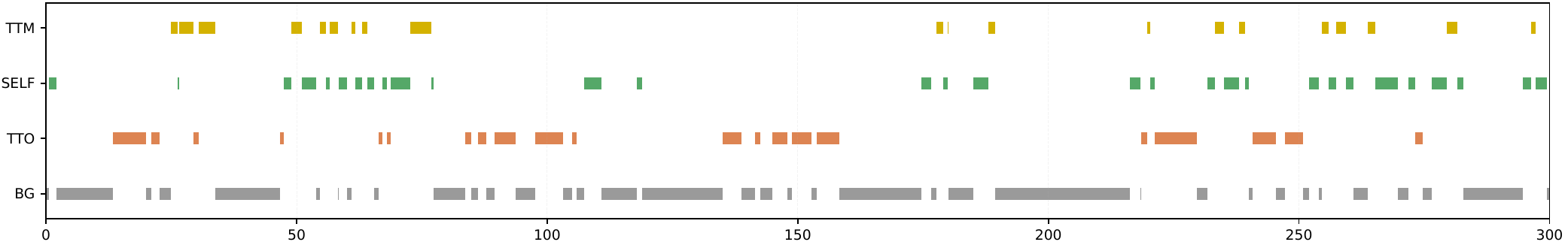}
        \vspace{-1.5mm}
        \centerline{\small (a)}

        \vspace{1mm}

        \makebox[\linewidth][l]{%
            \hspace*{0.028\linewidth}%
            \includegraphics[width=0.971\linewidth]{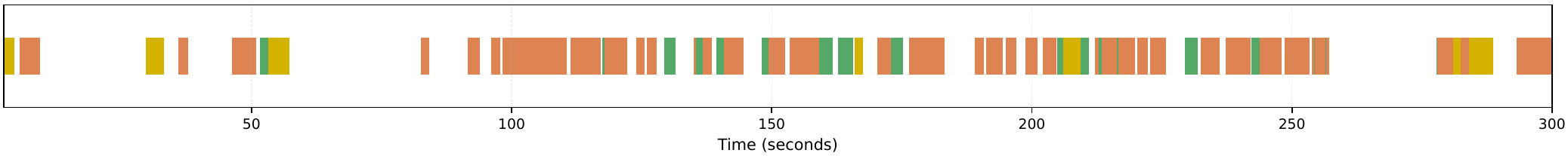}%
        }
        \vspace{-1.5mm}
        \centerline{\small (b)}
    \end{minipage}

    \vspace{-1mm}
    \caption{Qualitative illustration of the original Ego4D social
    annotations from two different clips. (a) One clip is dominated by
    long background intervals, while TTM segments are sparse and
    fragmented. (b) Another clip contains frequent transitions among
    speaking states, with blank intervals indicating frames without valid
    social-state annotations.}
    \label{fig:ego4d_annotation_examples}
\end{figure*}

\section{Related Work}

\subsection{Egocentric social interaction}

Egocentric social interaction understanding studies how surrounding people attend to and communicate with the camera wearer. A key step in making this problem measurable at scale is Ego4D, whose Social benchmark explicitly defines wearer-centered tasks such as Looking-at-Me (LAM) and Talking-to-Me (TTM) \cite{ego4d_social}. This benchmark shifts the focus from generic first-person scene understanding to a more interaction-centered question: whether a visible social partner is directing attention or speech toward the wearer.

Building on this benchmark, subsequent work has mainly advanced TTM within the original wearer-directed prediction setting. EgoTask Translation (EgoT2) shows that heterogeneous Ego4D tasks can benefit one another through cross-task translation, indicating that wearer-centered social prediction can leverage broader egocentric task structure \cite{xue2023egot2, peirone2024egopack}. QuAVF then targets the Ego4D TTM challenge more directly through quality-aware audio-visual fusion, showing that reliability-aware integration of visual and audio cues improves benchmark performance \cite{lin2023quavf}. Recent work has further expanded the problem from local audio-visual evidence to richer interaction structure \cite{huang2023egoaol, chen2024soundingactions}. In particular, long-term social context has been shown to be important for egocentric addressee detection \cite{kong2024long}, while AV-CONV moves beyond isolated wearer-directed decisions by modeling conversational relations among multiple participants from egocentric observations \cite{jia2024avconv}. Face-voice association and missing-modality robustness further broaden wearer-centered speech understanding beyond binary wearer-directed classification \cite{tao2024facevoice, qian2026egoadapt, saeed2024fame}.
CASTLE 2024 and its associated ACM Multimedia Grand Challenge
support multimodal analysis of synchronized ego- and exo-centric
recordings, but do not specifically address frame-level
wearer-directed speaking-state recognition
~\cite{rossetto2025castle,rossetto2025overview}.

\subsection{Active Speaker Detection}

Active Speaker Detection (ASD) asks whether a visible face is speaking at a given moment and has become a core problem in audio-visual scene understanding. AVA-ActiveSpeaker established a large-scale benchmark for this task and enabled systematic study of frame-level speaker activity from synchronized face tracks and audio \cite{roth2020ava}. With this benchmark in place, subsequent work progressively strengthened the modeling toolkit. Earlier ASD and meeting-oriented systems also explored weak supervision and meeting-specific audiovisual fusion \cite{chakravarty2016cotraining, madrigal2020meetingasd, pibre2023fusionmeetings, berghi2024arrayasd}. TalkNet highlights the importance of long-range temporal reasoning for audio-visual active speaker detection \cite{tao2021talknet}. EASEE pushes ASD toward an end-to-end formulation by jointly optimizing representation learning and speaker prediction \cite{alcazar2022easee}. ASD-Transformer shows that efficient self- and cross-modal transformer designs can be effective for ASD \cite{datta2022asdtransformer, radman2024asnet, wuerkaixi2022syncasd}. Target-specific formulations and richer fusion strategies have also been explored for harder speaker-selection settings \cite{jiang2023targetasd, xiong2022looklisten, assuncao2021biomodality, li2023biorepmm, yin2025audiofaces}. LoCoNet further combines long-term intra-speaker context with short-term inter-speaker competition, reflecting a more structured treatment of temporal and multi-person information in ASD \cite{wang2024loconet}.

More recent work extends ASD from offline recognition toward deployment-oriented settings. Gurvich et al.\ study an online causal ASD system designed for low-power edge use \cite{gurvich2023realtimeasd, liao2025lrasd}, and Kundu et al.\ analyze efficient and streaming ASD under explicit context constraints \cite{kundu2024streamingasd}. This evolution \cite{braga2024sslasd, tao2024mused} makes ASD highly relevant to our work from a methodological perspective, since it provides mature designs for multimodal fusion, temporal reasoning, and causal inference.
Standard ASD detects visible speakers, while egocentric TTM specifically identifies speech directed to the camera wearer ~\cite{yang2023avsd, kang2020multimodaldiarization, pan2024lateavfusion}. We adapt ASD methods to this distinct social setting by introducing a structured four-state prediction framework.

\begin{table*}[t]
\centering
\caption{Comparison between the Ego4D social annotations and our Online TTM Dataset.}
\label{tab:dataset_comparison}
\setlength{\tabcolsep}{5pt}
\renewcommand{\arraystretch}{1.15}
\small
\begin{tabular}{l l c c c c c l}
\toprule
Dataset & Prediction unit & Granularity & Videos & Total frames & TTM ratio & Online eval. & Benchmark purpose \\
\midrule
\textbf{Ego4D Social} 
& Segment label 
& Clip-level 
& 439 
& 3,951,000 
& 11.49\% 
& $\times$ 
& Offline clip recognition \\

\textbf{\shortstack[l]{Online TTM (ours)}} 
& State label 
& Frame-level 
& 406 
& 885,866 
& 41.49\% 
& \checkmark 
& Online frame prediction \\
\bottomrule
\end{tabular}
\end{table*}

\section{Problem Formulation and Dataset}

\subsection{Problem Definition}
We revisit talk-to-me (TTM) from the perspective of online egocentric social interaction and reformulate it as an online frame-level task. Given a first-person video stream, the goal is to infer, at each time step, whether a visible person is currently speaking to the camera wearer. Conventional TTM settings are typically organized around short video segments, where each segment is assigned a single speaking label, and interaction is treated as a coarse temporal unit. In contrast, we define TTM as a frame-level prediction task: at each time step, the model predicts the speaking state of the current moment rather than assigning a single label to an entire segment. This formulation provides a more direct characterization of online egocentric social interaction.

Formally, let $x_t$ denote the multimodal observation at time step $t$, including the visual scene and its associated audio context. Under the online setting, the model is allowed to access the current observation together with a finite history of past observations, but not future frames. Let $x_{1:t} = \{x_1, x_2, \dots, x_t\}$
denote the available observation sequence up to time step $t$. The model predicts
\begin{equation}
\hat{y}_t = f(x_{1:t}),
\end{equation}
where $\hat{y}_t$ is the estimated speaking state at the current moment.

Rather than treating TTM as a binary event against an undifferentiated negative class, we define the task over four speaking states:
\begin{equation}
y_t \in \{\text{BG}, \text{SELF}, \text{TTO}, \text{TTM}\},
\end{equation}
where BG denotes background or non-interactive conditions, SELF denotes self-talking by the wearer, TTO denotes talking-to-others, and TTM denotes talking-to-me.

Preserving this structure in the label space is more informative than collapsing all non-TTM cases into a single negative class. Although TTM remains the primary target, BG, SELF, and TTO provide socially distinct decision boundaries that help separate TTM from neighboring speaking states.

\subsection{Challenges of Online Egocentric TTM}
Online egocentric TTM remains challenging even under the structured four-state formulation. First, wearer-directed speech is often temporally local and dynamically evolving. In realistic first-person social interaction, speaking states may change within a short interval: a person may first talk to someone else and then turn to the wearer, and the wearer may begin speaking during an ongoing interaction. As a result, a key difficulty is not only whether TTM occurs, but also when it occurs and how it is separated from neighboring states over time.

Second, TTM must be distinguished from socially confusable speaking states rather than from a generic negative class. In particular, TTO and SELF often share strong visual and acoustic overlap with TTM. Talking-to-others still contains genuine speech and visible lip movements, while self-talking may also co-occur with salient audio activity and wearer-centered attention cues. These states therefore introduce ambiguity, making our task substantially more difficult than previous speech-versus-background detection.

Third, the task is subject to an online causal constraint. At inference time, the model must make a decision using only the current observation and its preceding context, without access to future frames. This prevents the model from relying on future evidence to resolve ambiguous moments and makes the problem more consistent with real deployment settings in wearable assistants, augmented reality systems, and interactive egocentric perception.

\begin{figure*}[t]
  \centering
  \includegraphics[width=0.8\linewidth]{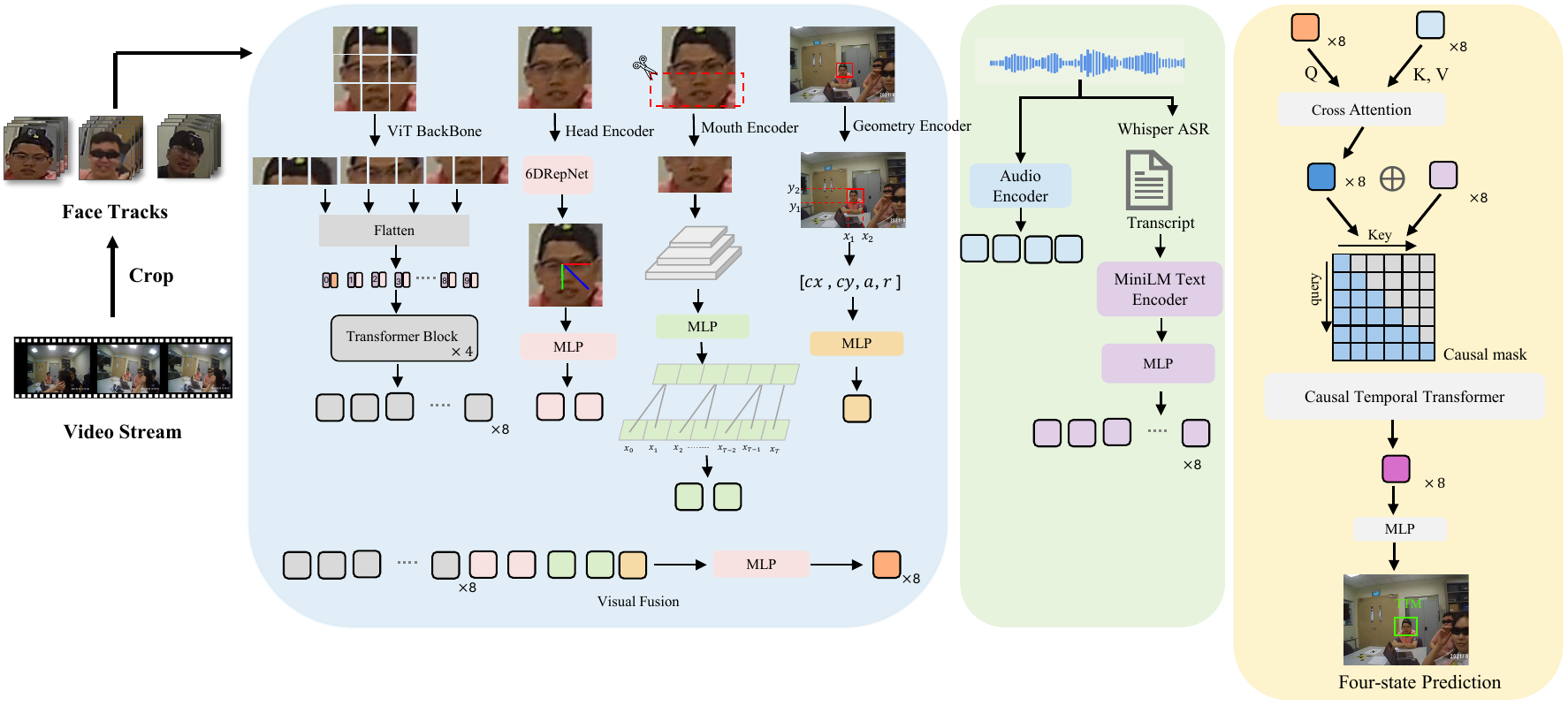}
  \caption{Overview of the proposed online multimodal model for egocentric social speaking recognition. From a continuous first-person video stream, we construct a person-centered visual stream and an aligned audio-text stream. The visual encoder extracts complementary cues from facial appearance, head-related representation, mouth motion, and wearer-centered geometry, while the audio-text branch captures frame-synchronous acoustic evidence and Whisper-derived speech semantics. These features are fused by cross-attention and then processed by a causal temporal Transformer with a causal mask to produce frame-level four-state predictions over BG, SELF, TTO, and TTM.}

   \Description{A pipeline diagram of the proposed online multimodal model. A continuous egocentric video stream is split into a person-centered visual branch and an aligned audio-text branch. The visual side contains modules for facial appearance, head-related features, mouth motion, and wearer-centered geometry. The audio-text side provides acoustic evidence and text-semantic features. These representations are fused through cross-attention and then sent to a causal temporal Transformer, which outputs frame-level predictions for four classes: background, self-talking, talking-to-others, and talking-to-me.}
  \label{fig:method_overview}
\end{figure*}
\subsection{Dataset Construction and Statistics}

The original Ego4D social annotations~\cite{ego4d_social} already include four speaking states, namely BG, SELF, TTO, and TTM. However, the annotations are organized for clip-level recognition, where each segment is assigned a single social-state label. While such a formulation is suitable for offline recognition, it is not directly applicable to the online frame-level TTM task studied in this paper, where the model must determine at each time step whether someone is currently speaking to the wearer. To address this mismatch, we first reconstruct frame-level supervision from the original segment-level annotations, so that each frame is associated with a speaking state. As illustrated in Fig.~\ref{fig:ego4d_annotation_examples}, the original Ego4D social annotations contain long background intervals, sparse TTM segments, frequent alternation among speaking states, and substantial unlabeled gaps.

Beyond the mismatch in temporal granularity, the original Ego4D distribution also exhibits a strong class imbalance: TTM accounts for only 11.49\% of all usable frames, while the remaining data are dominated by background, talking-to-others, and self-talking cases. To address this issue, we retain all usable TTM frames and re-sample BG, SELF, and TTO in a task-aware manner. Importantly, this re-sampling is not designed merely to achieve numerical balance. Instead, we preserve non-TTM samples that are temporally close to TTM events, locally continuous within interaction windows, and sufficiently informative around state boundaries. The resulting benchmark therefore not only alleviates the original class bias, but also provides a more structured negative space for distinguishing TTM from neighboring speaking states.

If the source data were sampled naively, many realistic state transitions would be lost, and the task would degenerate into isolated frame classification. To avoid this, we explicitly preserve a sufficient number of cross-state transitions during benchmark construction, so that the final data contain not only static speaking states, but also realistic transitions between them. In particular, around 40\% of the sampled windows contain at least two speaking states, allowing the benchmark to cover transitions such as TTO-to-TTM and SELF-to-BG. This design enables models to learn how one social speaking state changes into another, making the benchmark more faithful to the way wearer-directed speech emerges in online interaction and more suitable for analyzing why TTM is often confused with certain non-TTM states.

The original annotations also contain severe noise that can interfere with online modeling, including missing frames, incomplete annotation coverage, prolonged absence of valid face bounding boxes, and consecutive black-frame segments. If left untreated, such noise would introduce artificial discontinuities unrelated to the underlying interaction semantics. To improve data quality, we therefore apply a systematic cleaning procedure. For occasional face disappearance or local missing observations, we retain the natural perturbations that may occur in realistic scenes; for clearly corrupted sequences, we remove them. Specifically, we filter out samples with missing frames or abnormal annotation coverage, and discard sequences within the same labeled state when the proportion of black frames or no-face frames exceeds 40\%. Through this process, the resulting data preserve the complexity of realistic first-person interaction while substantially reducing severe noise that would otherwise affect model performance and analysis.

After frame-level reconstruction, task-aware re-sampling, and data cleaning, we obtain the Online TTM Dataset. The final benchmark contains 885,866 high-quality frame-level samples. Starting from 439 candidate videos, we retain 406 videos and split them into 334/21/51 videos for training, validation, and testing, respectively. The final label distribution comprises 367,529 TTM, 100,321 SELF, 170,321 TTO, and 247,695 BG frames, resulting in a substantially more balanced distribution than the original Ego4D annotations. More importantly, the Online TTM Dataset is not only a more suitable frame-level dataset in scale and balance, but also a benchmark that simultaneously supports online frame-level prediction, TTM-centric evaluation, and realistic socially complex interaction scenarios. A comparison between the original Ego4D social annotations and our Online TTM Dataset is summarized in Table~\ref{tab:dataset_comparison}.

\section{Method}

\subsection{Overview}

We propose a multimodal model for online egocentric TTM recognition. Given a short first-person temporal window, the model predicts the current social speaking state at each time step under a causal setting, where only the current and past observations are accessible. At time step $t$, the model receives a person-centered face crop $I_t$, a frame-synchronous audio input $x_t^{\mathrm{audio}}$, wearer-centered geometric cues $g_t$, a pre-extracted head representation $h_t$, and a text-semantic input $x_t^{\mathrm{text}}$. The model outputs a four-way prediction
\begin{equation}
\hat{y}_t \in \{\mathrm{BG}, \mathrm{SELF}, \mathrm{TTO}, \mathrm{TTM}\},
\end{equation}
corresponding to background, self-talking, talking-to-others, and talking-to-me, respectively.

As shown in Fig.~\ref{fig:method_overview}, the proposed framework consists of three components. First, a person-centered visual branch extracts complementary cues from target face tracks, including facial appearance, head-related information, mouth motion, and wearer-centered geometry. Second, an audio-text branch encodes frame-synchronous acoustic evidence together with Whisper-derived speech semantics. Third, the visual and audio-text representations are fused by cross-attention and then modeled by a causal temporal Transformer with a causal mask for online temporal reasoning, followed by frame-level four-state prediction.

\subsection{Visual Encoder}

The visual encoder captures complementary person-centered cues for online TTM, including facial appearance, head-related information, mouth dynamics, and wearer-centered geometry. For each frame, we use a target face crop as the primary visual input. When the annotated interaction target is available, we use the corresponding face bounding box; otherwise, we fall back to the largest visible face in the frame. For SELF frames, we use a zero-filled face crop together with zero-valued person-centered features.

\vspace{1mm}
\noindent
\textbf{Global facial feature extraction.}
We use a lightweight ViT-style encoder~\cite{dosovitskiy2021vit} to extract a global facial representation $v_t^{\mathrm{vit}}$ from each face crop in the target face track. Each face crop is resized to $224\times224$, tokenized into non-overlapping $16\times16$ patches, and processed by a 4-layer Transformer encoder with 4 attention heads. The final CLS token is used as the global facial representation. This branch encodes facial appearance and contextual information, providing a stable visual basis for subsequent fine-grained branches.

\vspace{1mm}
\noindent
\textbf{Head-focused representation.}
Global appearance alone is insufficient for determining whether the interaction is directed toward the wearer, since a visible person may not actually be facing the wearer while speaking (as shown in Sec.~\ref{subsec:ablation}). We therefore design a head-related branch. Specifically, we use a pretrained \textbf{6DRepNet}~\cite{10477888} to extract a 2048D hidden-layer representation for each face and feed this representation into a lightweight MLP to obtain a head feature $v_t^{\mathrm{head}}$. Compared with using only a few explicit pose angles, this design preserves richer information about head orientation and pose variation, such as whether the face is frontal, noticeably turned away, or undergoing motion such as turning or nodding.

\vspace{1mm}
\noindent
\textbf{Mouth-motion encoder.}
To explicitly model speaking activity, we design a mouth-motion branch. This branch focuses on the lower half of the face as an approximate mouth region, first extracts a compact static mouth representation with a small CNN, then computes temporal differences between adjacent frames, and finally applies causal temporal convolution to capture short-term lip-motion patterns. The motivation is that, if a person is speaking to the wearer, the mouth should exhibit temporally evolving motion rather than remain visually static. Temporal differencing encourages the model to focus on changes in mouth configuration instead of static mouth appearance, while causal temporal convolution further strengthens the modeling of local articulation dynamics.

\vspace{1mm}
\noindent
\textbf{Geometry encoder.}
Spatial position is also an important cue for identifying the interaction target in egocentric video, especially when multiple faces appear in the same frame. We therefore derive wearer-centered geometric cues from the selected face bounding box and represent them as $g_t = [c_x, c_y, a, r],$
where $c_x$ and $c_y$ denote the normalized box center, $a$ is the normalized box area, and $r$ is the aspect ratio. We then project $g_t$ into a learned embedding $v_t^{\mathrm{geo}}$. These cues provide an explicit spatial prior: the person speaking to the wearer is often more central, visually larger, and more prominent in the first-person view, whereas small or peripheral faces are more likely to be side participants or background people. This wearer-centered geometry is particularly helpful for disambiguation in multi-person scenes.

Finally, we concatenate all branches into a unified visual representation $v_t = [v_t^{\mathrm{vit}}; v_t^{\mathrm{head}}; v_t^{\mathrm{mouth}};  v_t^{\mathrm{geo}}].$
This representation jointly encodes who the visible person is, whether the person is speaking, whether the head is oriented toward the wearer, and whether the face is spatially interaction-relevant, making it better aligned with the evidence required for our online egocentric TTM task.

\subsection{Audio and Speech-Semantic Encoding}

Acoustic evidence alone is often insufficient for socially grounded TTM recognition, especially when TTM and TTO share similar speech patterns and background noise. To address this issue, we jointly model acoustic evidence and speech semantics. At each time step, we extract a frame-synchronous waveform chunk from the original audio stream and align an utterance-level semantic feature to the current video frame according to its temporal span. The waveform chunk is then encoded with a lightweight temporal CNN to produce an audio representation. This branch captures local speech-related acoustic patterns and provides direct evidence of speaking activity, complementing the visual cues from the person-centered stream.

However, the social meaning of an utterance is often reflected more strongly in its semantic content than in acoustics alone. We therefore further incorporate Whisper~\cite{radford2023robust} to extract speech-semantic representations derived from ASR transcripts, aligned with the temporal span of each utterance. These representations are projected into the shared multimodal space and serve as an additional semantic cue. The intuition is that, in conversational settings, spoken content may implicitly indicate whether an utterance is directed toward the wearer, participates in the current interaction, or corresponds to background speech. Compared with raw acoustic signals, such semantic information can provide stronger cues for socially grounded TTM detection.
The audio-text branch can be summarized as
\begin{equation}
a_t = E_{\mathrm{aud}}(x_t^{\mathrm{audio}}), \qquad
\tilde{s}_t = x_t^{\mathrm{text}},
\end{equation}
where $a_t$ denotes the acoustic representation and $\tilde{s}_t$ denotes the pre-extracted text-semantic feature. In implementation, we first use Whisper to perform automatic speech recognition (ASR) and generate speech transcripts, and then apply a MiniLM text encoder~\cite{wang2020minilm} to extract a text-semantic feature from transcripts.

\begin{table*}[t]
\centering
\caption{Comparison with baselines on the Online TTM Dataset.}
\label{tab:main_comparison}
\setlength{\tabcolsep}{4pt}
\renewcommand{\arraystretch}{1.10}
\small
\scalebox{0.8}{
\begin{tabular}{lcccc ccc}
\toprule
\multicolumn{1}{c}{ } & \multicolumn{4}{c}{Frame-level F1 (\%)} & \multicolumn{1}{c}{Segment-level (\%)} & \multicolumn{2}{c}{Efficiency} \\
\cmidrule(r){2-5} \cmidrule(lr){6-6} \cmidrule(l){7-8}
Method & BG & SELF & TTO & TTM & TTM mAP & Params (M) & Model latency (ms/frame) \\
\midrule
Ego4D baseline \cite{ego4d_social} & 49.9 & 59.2 & 2.3  & 61.3 & 62.6 & 18.066 & 8.2 \\
Streaming ASD \cite{kundu2024streamingasd}  & 66.5 & 69.4 & 25.3 & 66.3 & 72.8 & 3.451  & 4.0 \\
CMeRT \cite{pang2025cmert}          & 64.9 & 71.0 & 25.0 & 65.9 & 71.0 & 38.088 & 12.0 \\
LoCoNet \cite{wang2024loconet}        & 64.0 & 63.4 & 22.7 & 63.3 & 70.6 & 17.304 & 7.4 \\
ProTAS~\cite{Shen_2024_CVPR}
& 62.3 & 60.7 & 20.3 & 60.1 & 68.5 & 20.638 & 6.4 \\

\midrule
\textbf{Ours}  & \textbf{99.4} & \textbf{72.6} & \textbf{40.2} & \textbf{75.5} & \textbf{79.8} & \textbf{5.876} & \textbf{4.49} \\
\bottomrule
\end{tabular}
}
\end{table*}

\subsection{Online Multimodal Fusion and Causal Temporal Modeling}

Since social speaking states evolve over time, the model must jointly perform multimodal fusion and causal temporal modeling under an online constraint. We therefore first project the visual and audio features and fuse them via cross-attention in a shared hidden space. Specifically, visual features are used as queries, while audio features are used as keys and values, so that speech-related evidence can refine the person-centered visual stream. The same windowed causal mask is also applied in cross-attention, ensuring that cross-modal interaction at time step $t$ depends only on the current and recent past observations. We also explored the reverse direction, but found it to be noisier in practice; therefore, we adopt visual-query, audio-value fusion in the final model.

After cross-modal fusion, the projected text-semantic feature is added to the fused representation. This allows the model to integrate not only visual appearance and acoustic activity, but also higher-level semantic context associated with speech. The resulting multimodal sequence is then passed to a causal temporal Transformer, where each block consists of masked multi-head self-attention and a feed-forward network. To enforce online inference, we apply a windowed causal attention mask
\begin{equation}
M_{ij} =
\begin{cases}
0, & i-\tau+1 \le j \le i,\\
-\infty, & \text{otherwise},
\end{cases}
\end{equation}
where $\tau = 32$ is the past context length. This mask ensures that the token at time step $i$ only attends to the current and recent past observations, without access to future frames.

The overall online fusion-and-temporal reasoning process can be summarized as

\begin{equation}
p_t=\mathrm{Softmax}\!\left(
W\left[
E_{\mathrm{temp}}\!\left(
\mathrm{CrossAttn}(v_{1:T},a_{1:T};M)+W_s \tilde{s}_{1:T};\,M
\right)
\right]_t
\right),
\end{equation}
where $W_s$ denotes a learnable projection for the speech-semantic feature, $E_{\mathrm{temp}}$ denotes the causal temporal Transformer, and $M$ is the windowed causal attention mask.

\subsection{Training Objective}

Under the four-state formulation, given the predicted class distribution $p_t$ at time step $t$, we optimize the model using frame-level cross-entropy over all valid, non-padded frames:
\begin{equation}
\mathcal{L} = - \sum_{t \in \Omega} \sum_{c=1}^{4} y_{t,c}\log p_{t,c},
\end{equation}
where $\Omega$ denotes the set of valid time steps in a batch. In practice, padded positions are ignored in both the loss computation and the attention masks, and label smoothing is applied during training for more stable optimization.

\section{Experiments}

\subsection{Experimental Setup}

All experiments are conducted on the proposed Online TTM Dataset. For fair comparison, all baselines are adapted to the same four-state label space and causal online setting, and are retrained under the same train/val/test split.
At the frame level, we report F1 for each class, with particular emphasis on TTM F1 in the main results. At the segment level, we report TTM segment mAP.
All models are implemented in PyTorch. Training is performed using AdamW with an initial learning rate of $3\times10^{-4}$, a batch size of 512, and 60 training epochs. We further report model size in Params (M) and model-only inference
latency in ms/frame as efficiency indicators.

\begin{figure*}[t]
  \centering
  \includegraphics[width=0.9\linewidth]{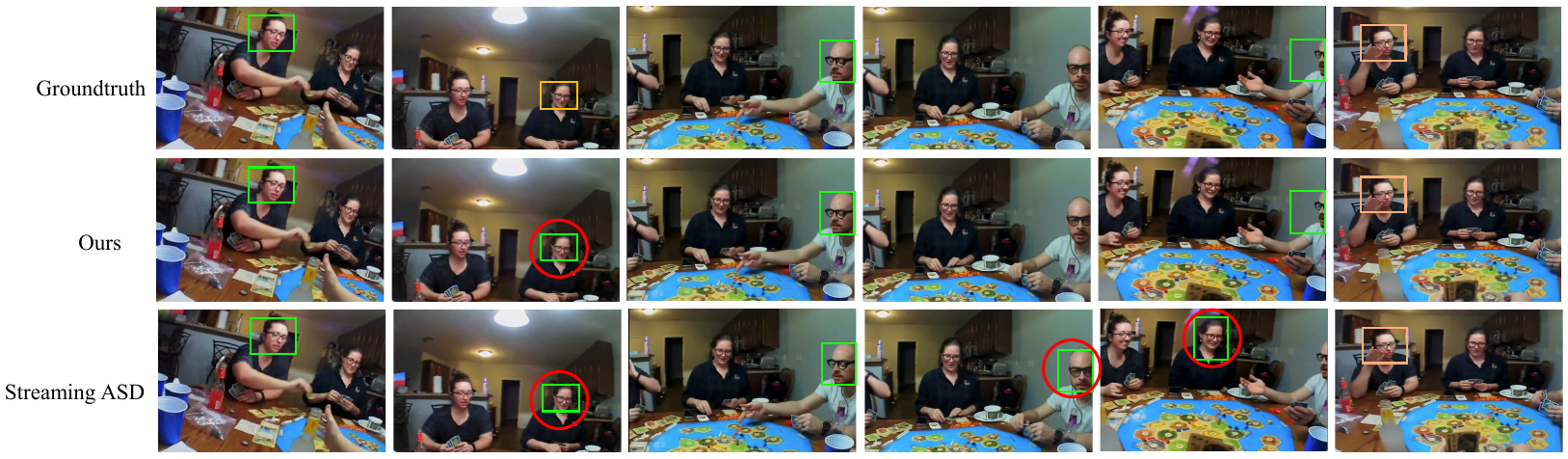}
    \caption{\textbf{Results comparison of our method and Streaming ASD on challenging scenarios from the Online TTM test set.} \textcolor{green!60!black}{Green boxes} denote TTM targets, \textcolor{orange!90!black}{orange boxes} denote TTO targets, and \textcolor{red}{red circles} indicate false predictions. The first row shows the ground truth, the second row shows our predictions, and the third row shows the predictions of Streaming ASD.}
    
    \Description{A qualitative comparison figure showing several challenging examples from the Online TTM test set. Each example is arranged in three rows: the first row shows the ground truth, the second row shows predictions from the proposed method, and the third row shows predictions from Streaming ASD. Faces identified as talking-to-me or talking-to-others are marked with colored boxes, and incorrect predictions are marked with circles. The figure shows that the proposed method more often places the correct target annotation on socially confusable faces than the baseline.}

  \label{fig:TTM Result}
\end{figure*}
\subsection{Experimental Results}

We compare our method against several representative baselines under the same online four-state protocol. The Ego4D baseline \cite{ego4d_social} serves as a task-specific reference inherited from the original benchmark. Streaming ASD \cite{kundu2024streamingasd} serves as a causal audio-visual streaming baseline under strict online inference constraints. CMeRT \cite{pang2025cmert} and LoCoNet \cite{wang2024loconet} provide two representative temporal backbones for online context modeling. ProTAS~\cite{Shen_2024_CVPR} is adapted as an additional
progress-aware causal temporal baseline.

Table~\ref{tab:main_comparison} reports the main comparison results on the Online TTM Dataset. Our method achieves the best results on all four class-wise F1 metrics and the highest TTM segment mAP, including 75.5\% frame-level F1 on TTM and 79.8\% TTM segment mAP. Table~\ref{tab:main_comparison} also compares model size and inference latency. Our method achieves the best overall recognition performance while
remaining computationally efficient, with a compact model size of
5.876M parameters and a model-only inference latency of
4.49 ms/frame. Including face detection, 6DRepNet, Whisper ASR,
and MiniLM, the complete streaming pipeline runs at 48.1 ms/frame
(20.8 FPS), with Whisper accounting for most of the runtime.

We further analyze the effects of the label formulation and temporal
setting. Four-state training improves TTM F1 from 64.2\% under binary
training to 75.5\%, indicating that explicitly modeling socially
distinct negative states provides more informative decision boundaries.
Offline inference achieves 87.2\% TTM F1, compared with 75.5\% under
the causal online setting, quantifying the performance cost of operating
without future observations.

Compared with Streaming ASD, our method improves TTM F1 by 9.2\% and TTM segment mAP by 7.0\%. The gain on TTO is even larger, reaching 14.9\%, which is particularly important because TTO is one of the most socially confusable competing states for TTM. These results suggest that our method better distinguishes wearer-directed speech from speech directed to others. Compared with the Ego4D baseline, the improvement is even larger across all socially meaningful speaking states, indicating that simply adapting a generic benchmark baseline to the online frame-level setting is insufficient. Figure~\ref{fig:TTM Result} further presents qualitative comparisons with Streaming ASD on challenging examples from the Online TTM test set, where our method produces more accurate predictions in socially confusable cases.

\begin{table}[t]
\centering
\caption{Ablation study of the main components.}
\label{tab:ablation_progressive}
\setlength{\tabcolsep}{1.6pt}
\renewcommand{\arraystretch}{1.15}
\small
\scalebox{0.8}{
\begin{tabular}{ccccc cccc c}
\toprule
\multicolumn{5}{c}{Components} & \multicolumn{4}{c}{Frame-level F1 (\%)} & \multicolumn{1}{c}{Segment-level (\%)} \\
\cmidrule(r){1-5} \cmidrule(lr){6-9} \cmidrule(l){10-10}
ViT & Head & Mouth & Geom & Text & BG & SELF & TTO & TTM & TTM mAP \\
\midrule
\cmark &        &        &        &        & 56.33 & 14.28 & 0.05  & 62.27 & 67.00 \\
\cmark & \cmark &        &        &        & 64.80 & 63.21 & 0.15  & 63.23 & 71.07 \\
\cmark & \cmark & \cmark &        &        & 66.68 & 70.45 & 19.42  & 64.39 & 72.13 \\
\cmark & \cmark & \cmark & \cmark &        & 70.67 & 70.88 & 26.07  & 65.00 & 72.21 \\
\cmark & \cmark & \cmark & \cmark & \cmark & \textbf{99.42} & \textbf{72.60} & \textbf{40.19} & \textbf{75.54} & \textbf{79.78} \\
\bottomrule
\end{tabular}
}
\end{table}

\subsection{Ablation}
\label{subsec:ablation}

We next analyze the contribution of the main components. Starting from a ViT-only baseline, we gradually add the Head-focused representation (Head), the Mouth-motion encoder (Mouth), the Geometry encoder (Geom), and finally the Whisper-derived text-semantic branch (Text). The results are summarized in Table~\ref{tab:ablation_progressive}.

First, adding Head on top of ViT leads to clear improvements on BG, SELF, and TTM, indicating that head-related cues provide useful information for distinguishing socially meaningful speaking states. This is consistent with the intuition that wearer-directed interaction is closely related to whether a visible person is oriented toward the wearer. Second, adding Mouth further improves the results, especially on SELF and TTO, suggesting that local mouth dynamics provide useful evidence for speaking activity, although their gain on TTM is relatively limited. Third, introducing Geom yields additional improvements, particularly on BG and TTO, supporting our hypothesis that wearer-centered spatial priors, such as face centrality and prominence in the egocentric view, are important for identifying socially relevant targets.

The largest gain comes from the introduction of the text-semantic branch. When Whisper-derived text-semantic features are added, performance improves substantially across all classes, with especially large gains on BG, TTO, and TTM. In particular, the frame-level F1 of TTM increases from 65.00\% to 75.54\%, while the TTM segment mAP rises from 72.21\% to 79.78\%. The improvement on TTO is also substantial, increasing from 26.07\% to 40.19\%. These results suggest that although progressively enriched visual cues improve social speaking recognition, higher-level semantic information is critical for resolving cases where audio-visual evidence alone remains ambiguous.

Additional robustness tests show graceful degradation under
missing or corrupted inputs. Zeroing the audio or face input yields
72.8\% and 69.4\% TTM F1, respectively, while 30\% and 50\%
face-box dropout result in 69.8\% and 67.2\% TTM F1.
Together with the no-text result in Table~\ref{tab:ablation_progressive},
these results indicate that the model does not rely exclusively on
any single modality.

To examine whether the text-semantic branch mainly exploits
transcript-level shortcuts, we further shuffle transcripts across
speaking segments while preserving silent intervals. Under this
control, BG F1 remains nearly unchanged, decreasing from 99.42\%
to 98.8\%, whereas TTM F1 drops from 75.5\% to 69.6\%.
This result suggests that the BG improvement is largely associated
with speech-presence information, while accurate TTM recognition
benefits from temporally aligned semantic content.

\begin{table}[t]
\centering
\caption{TTM-centered confusion attribution with and without text
semantics. All values denote the number of frames.}
\label{tab:error_attribution}
\setlength{\tabcolsep}{5.5pt}
\renewcommand{\arraystretch}{1.15}
\small
\scalebox{0.8}{
\begin{tabular}{lccc ccc}
\toprule
& \multicolumn{3}{c}{TTM False Negatives}
& \multicolumn{3}{c}{TTM False Positives} \\
\cmidrule(r){2-4} \cmidrule(l){5-7}
Variant
& BG & SELF & TTO
& BG & SELF & TTO \\
\midrule
No Text
& 17,567 & 9,511 & 49
& 15,583 & 1,305 & 25,694 \\

\textbf{Full Model}
& \textbf{111} & \textbf{5,106} & \textbf{14,187}
& \textbf{0} & \textbf{5,297} & \textbf{21,493} \\
\bottomrule
\end{tabular}
}
\end{table}
\subsection{Error Attribution Analysis}
\label{subsec:error_analysis}

The text-semantic branch produces the largest redistribution of
TTM-related errors among all model components. We therefore focus on
the TTM-centered confusion patterns of the full model and the variant
without text semantics. Specifically, we analyze:
(1) \textbf{false negatives}, where the ground-truth label is TTM but
the model predicts a non-TTM state; and
(2) \textbf{false positives}, where the model predicts TTM but the
GT label belongs to another state.

As shown in Table~\ref{tab:error_attribution}, removing the
text-semantic branch causes substantial confusion between BG and TTM.
Aggregating both directions, BG-related TTM confusion decreases from
33,150 frames without text semantics to only 111 frames in the full
model. This shows that semantic information largely resolves
speech-versus-background ambiguity.

After adding text semantics, the largest remaining error is TTO
predicted as TTM, with 21,493 frames, followed by TTM predicted as
TTO, with 14,187 frames. The remaining errors therefore concentrate
on socially similar speaking states rather than background conditions.
This supports our task formulation: the main challenge is not merely
detecting whether speech is present, but determining whether the speech
is directed to the camera wearer.

\section{Conclusion}
We revisit egocentric talk-to-me (TTM) recognition from the perspective of online social interaction. Instead of the conventional offline clip-level formulation, we reformulate TTM as an online frame-level prediction task under a causal setting, enabling moment-level reasoning about wearer-directed speech. To support this setting, we construct the Online TTM Dataset, a large-scale benchmark derived from Ego4D Social with approximately 900K frame-level annotations, preserving realistic transitions among background, self-talking, talking-to-others, and talking-to-me states. We further develop an online multimodal model that integrates visual cues, wearer-centered geometry, acoustic features, and ASR-derived speech semantics, achieving 75.5\% frame-level F1 on TTM and outperforming strong baselines.

\begin{acks}
This work was supported by the National Eye Institute of the National Institutes of Health (NIH) under Award Number R01EY037100. The content is solely the responsibility of the authors and does not necessarily represent the official
views of NIH.
\end{acks}

\bibliographystyle{ACM-Reference-Format}
\bibliography{refs.bib}




\end{document}